\documentclass[a4paper, conference]{IEEEtran}

\usepackage[a4paper,
            top=54pt,
            bottom=114pt,
            left=37pt,
            right=38pt]{geometry}
            
\IEEEoverridecommandlockouts

\usepackage[utf8]{inputenc} 
\usepackage[T1]{fontenc}    
\usepackage{hyperref}       
\usepackage{url}            
\usepackage{booktabs}       
\usepackage{amsfonts}       
\usepackage{amssymb}
\usepackage{nicefrac}       
\usepackage{microtype}      
\usepackage{lipsum}
\usepackage{graphicx}
\usepackage{enumitem}       

\usepackage{tikz}
\usepackage{pgfplots}
\pgfplotsset{compat=1.3}
\usepackage{pgfplotstable}
\usepackage{textcomp}
\usetikzlibrary{shapes,arrows}

\tikzset{%
  block/.style    = {draw, thick, rectangle, minimum height = 3em,
    minimum width = 3em},
  sum/.style      = {draw, circle, node distance = 2cm}, 
  input/.style    = {coordinate}, 
  output/.style   = {coordinate} 
}

\newcommand{\tx}{\mathbf{x}}
\newcommand{\figurewidth}{0.49\textwidth}

\begin{document}

\title{ThriftyNets : Convolutional Neural Networks with Tiny Parameter Budget}

\author{

\IEEEauthorblockN{Guillaume Coiffier}
\textit{ENS de Lyon}\\
Lyon, France \\
\texttt{guillaume.coiffier@ens-lyon.fr}

\and

\IEEEauthorblockN{Ghouthi Boukli Hacene}
\textit{MILA}\\
Montréal, Canada\\

\and

\IEEEauthorblockN{Vincent Gripon}
\textit{IMT Atlantique and Université Côte d'Azur}\\
Brest, France\\

}

\maketitle

\begin{abstract}

Typical deep convolutional architectures present an increasing number of feature maps as we go deeper in the network, whereas spatial resolution of inputs is decreased through downsampling operations. This means that most of the parameters lay in the final layers, while a large portion of the computations are performed by a small fraction of the total parameters in the first layers.
In an effort to use every parameter of a network at its maximum, we propose a new convolutional neural network architecture, called ThriftyNet. In ThriftyNet, only one convolutional layer is defined and used recursively, leading to a maximal parameter factorization. In complement, normalization, non-linearities, downsamplings and shortcut ensure sufficient expressivity of the model.
ThriftyNet achieves competitive performance on a tiny parameters budget, exceeding 91\% accuracy on CIFAR-10 with less than 40K parameters in total, and 74.3\% on CIFAR-100 with less than 600K parameters.

\end{abstract}


\section{Introduction}

Convolutional Neural Networks (CNNs) have revolutionized the field of computer vision, providing consistent state of the art results in a wide range of tasks, from image recognition to semantic segmentation. This increase in performance was accompanied with an increase in the depth, size and overall complexity of the corresponding models. It is not unusual to encounter models with hundreds of layers and tens of millions of parameters. This is even more true in the field of Natural Language Processing where very large models usually lead to the best performance.

There are multiple reasons why it would be desirable to reduce the size of CNNs. For example, for some applications, it is required to deploy systems on resource-constrained hardware (e.g. edge systems) or to provide real time predictions (e.g. assisted surgery, autonomous vehicles\dots). More generally, deep learning systems are often deployed as black boxes trained on huge available datasets, and are therefore lacking understandability and interpretability: reducing the number of parameters can help in visualizing and explaining decisions. Additionally, datacenters are increasingly used for deep learning, and their impact on the environment is becoming a concern.

For all these reasons, there have been a significant amount of works aiming at reducing the size and computational cost of CNNs. To cite a few, some works propose to prune the connections in deep learning systems so as to reduce their numbers~\cite{han2015learning}. Other works propose to distillate the knowledge from large architecture to smaller ones~\cite{hinton2015distilling}. It is also possible to focus on reducing the bit precision of weights and/or activations~\cite{courbariaux2015binaryconnect} or to factorize some of the operations~\cite{wu2018deep, han2015deep}. Finally, a lot of efforts have been dedicated to finding efficient architectures~\cite{howard2017mobilenets, iandola2016squeezenet, tan2019efficientnet}.

A key difficulty in the field is tied to the fact that there are multiple metrics to act upon, and a lot of possible hardware targets, on which some methods might not be applicable. Throughput, latency, energy, power, flexibility and scalability are among the most discussed ones in the literature. In this work, we focus on reducing the \emph{number of parameters} of architectures, which is usually strongly connected to the memory usage of the model. In this area, factorizing methods, which identify similar sets of parameters and merge them~\cite{wu2018deep}, are particularly effective, in that they considerably reduce the number of parameters while maintaining the same global structure and number of flops. Our contribution can be thought of as a factorization technique.

In this work, we propose to introduce a new factorized deep learning model, in which the factorization is not learned during training, but rather imposed at the creation of the model. We call these models \emph{ThriftyNets}, as they typically contain a very constrained number of parameters, while achieving top-tier results on standard classification vision datasets. The core idea we introduce consists in recycling the same convolution to be applied a large number of times when processing an input element. In more details, the main claims of our paper are:

\begin{enumerate}
    \item We introduce ThriftyNets, a new family of deep learning models that are designed with a fixed number of parameters and variable depths and flops.
    \item We perform experiments on standard vision datasets and show that we are able to outperform current deep learning solutions for tiny parameters budget.
    \item We design experiments to stress the impact of the hyperparameters of the proposed models on the accuracy and the flops of obtained solutions.
\end{enumerate}

The outline of the paper is as follows. In Section~\ref{sec:rw} we present related work and discuss the context of our proposed method. In Section~\ref{sec:methodo} we introduce the proposed methodology and discuss the role of hyperparameters on the total number of parameters. In Section~\ref{sec:expe} we perform experiments using standard vision datasets and compare the proposed method with existing alternative architectures. Section~\ref{sec:conclusion} is a conclusion.

\section{Related Work}
\label{sec:rw}

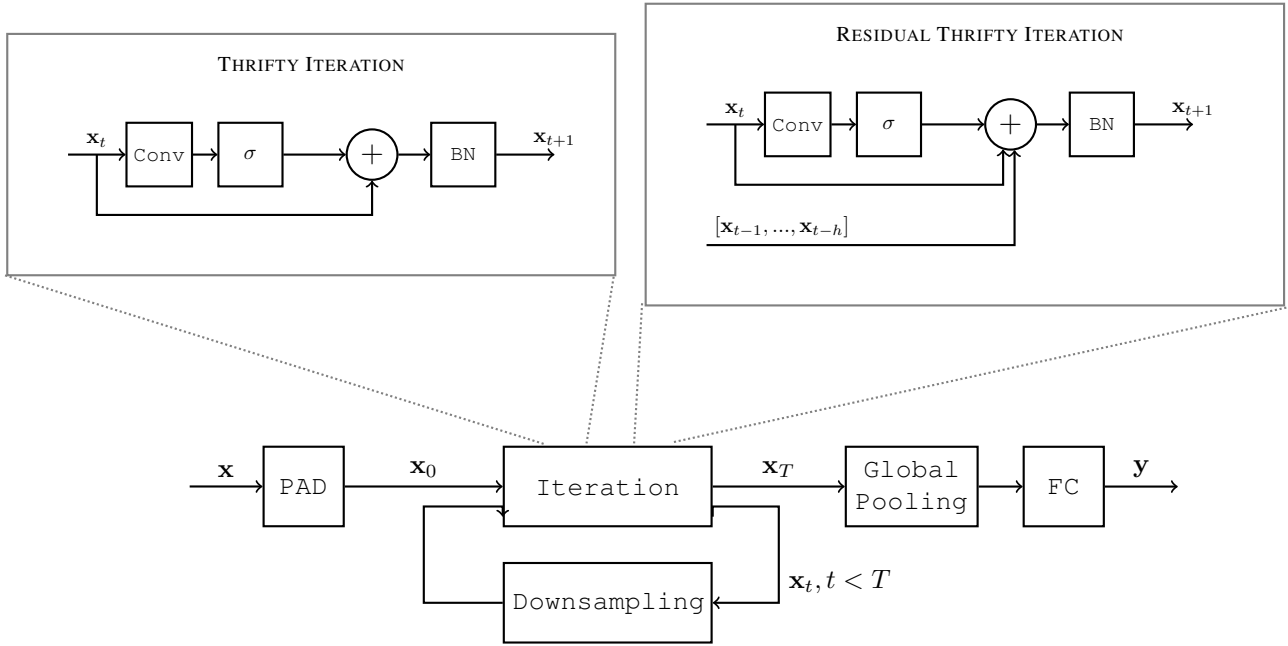
\begin{figure*}[t]
    \centering
    \begin{tikzpicture}
    [auto,
    thick,
    node distance=1.5cm,
    scale=0.8,
    every node/.style={scale=0.8}]
    
    \node [name=Aanchor] at (-1,-0.5) {};
    \draw [color=gray,thick](-2 , -2.5) rectangle (8, 1.5);
    
    \node [name=Banchor] at (9.5,0) {};
    \draw [color=gray,thick](8.5,-3) rectangle (19,2);
    
    \node [name=Canchor] at (1,-6) {};
    
    \node at (3,1) {\textsc{Thrifty Iteration}};
    
    \node [input, name=Ainput, below of=Aanchor, node distance=0cm] {};
    \node [block, name=Aconv, right of=Ainput] {\texttt{Conv}};
    \node [block, name=Aactiv, right of=Aconv] {$\sigma$};
    \node [sum, name=Asum, right of=Aactiv] {\Large$+$};
    
    \node [block, name=Abn, right of=Asum] {\texttt{BN}};
    \node [input, name=Aoutput, right of=Abn] {};

    \draw [->] (Ainput) -- node[name=Abranch] {$\tx_{t}$} (Aconv);
    \draw [->] (Aconv) --  (Aactiv);
    \draw [->] (Aactiv) -- (Asum);
    \draw [->] (Asum) -- (Abn);
    \draw [->] (Abn) -- node[at end] {$\tx_{t+1}$} (Aoutput);
    
    \coordinate [name=Ashortcut, below of=Aconv, node distance=1cm] {};
    \draw [->] (Abranch) |- (Ashortcut) -| (Asum);
    
	\node at (14,1.5) {\textsc{Residual Thrifty Iteration}};
	
    \node [input, name=Binput, below of=Banchor, node distance=0cm] {};
    \node [input, name=Binput3, below of=Binput, node distance=2cm] {};
    
    \node [block, name=Bconv, right of=Binput] {\texttt{Conv}};
    \node [block, name=Bactiv, right of=Bconv] {$\sigma$};
    \node [sum, name=Bsum, right of=Bactiv] {\Large$+$};
    
    \node [block, name=Bbn, right of=Bsum] {\texttt{BN}};
    \node [input, name=Boutput, right of=Bbn] {};
    \node [input, name=Boutput2, below of=Boutput, node distance=1cm] {};
    \node [input, name=Boutput3, below of=Boutput2, node distance=1cm] {};

    \draw [->] (Binput) -- node[name=Bbranch] {$\mathbf{x}_{t}$} (Bconv);
    \draw [->] (Bconv) --  (Bactiv);
    \draw [->] (Bactiv) -- (Bsum);
    \draw [->] (Bsum) -- (Bbn);
    \draw [->] (Bbn) -- node[at end] {$\tx_{t+1}$} (Boutput);
    
    \coordinate [name=Bsum2, below of=Bsum, node distance=1cm] {};
    \coordinate [name=Bsum3, below of=Bsum, node distance=2cm] {};
    
    \draw [->] (Bbranch) |- ([xshift=-0.3cm]Bsum2) -| ([xshift=-0.3cm]Bsum);
    
    \draw [->] (Binput3) -- node[at start, above right] {$[\tx_{t-1}, ..., \tx_{t-h}]$} (Bsum3) -| ([xshift=0.3cm]Bsum);
    
    \tikzset{
        scale=1.0,
        every node/.style={scale=1.0}
    }
    
    \node [input, name=Cinput, below of=Canchor, node distance=0cm] {};
    \node [block, name=Cpad, right of=Cinput] {\texttt{PAD}};
    \node [block, name=Cit, right of=Cpad, node distance=4cm, align=center, text width=2.5cm] {\texttt{Iteration}};
    \node [block, name=Cds, below of=Cit, anchor=north, text width=2.5cm, align=center, node distance=1cm] {\texttt{Downsampling}};
    \node [block, name=Cgmp, right of=Cit, node distance=4cm, text width=1.5cm, align=center] {\texttt{Global Pooling}};
    \node [block, name=Cfc, right of=Cgmp, node distance=2cm] {\texttt{FC}};
    \node [input, name=Coutput, right of=Cfc] {};

    \draw [->] (Cinput) -- node {$\tx$} (Cpad);
    \draw [->] (Cpad) -- node[name=Cbit2] {$\tx_0$} (Cit);
    \draw [->] (Cit) -- node[name=Cait2] {$\tx_T$} (Cgmp);
    \draw [->] (Cgmp) -- (Cfc);
    \draw [->] (Cfc) -- node {$\mathbf{y}$} (Coutput);
    
    \coordinate [name=Cbit, below of=Cbit2, node distance=0.1cm]{};
    \coordinate [name=Cait, below of=Cait2, node distance=0.1cm]{};
    
    \draw [->] ([yshift=-0.5cm]Cit.east) |- ([yshift=-0.5cm]Cait) |- node[above right] {$\mathbf{x}_t, t<T$} (Cds);
    \draw [->] (Cds) -| ([yshift=-0.5cm]Cbit) -| ([yshift=-0.5cm]Cit.west);

    
    \node [name=A1] at (-2.2, -2.45) {};
    \node [name=A2] at (8, -2.36) {};
    
    \node [name=B1] at (8.45, -2.8) {};
    \node [name=B2] at (19.2, -3) {};
    
    \node [name=C1] at (7, -5.35) {};
    \node [name=C2] at (7.5, -5.45) {};
    \node [name=C3] at (8.3, -5.45) {};
    \node [name=C4] at (8.8, -5.3) {};
    
    \draw [color=gray, densely dotted] (A1) -- (C1);
    \draw [color=gray, densely dotted] (A2) -- (C2);
    \draw [color=gray, densely dotted] (B1) -- (C3);
    \draw [color=gray, densely dotted] (B2) -- (C4);

\end{tikzpicture}
    \caption{Flow diagram of our algorithm. The typical three-channeled input is first padded with zeros to match a predetermined number of filters. Then, ThriftyNet performs a user-defined amount of iterations $T$, consisting of a convolution with the filter, a non-linear activation, a shortcut and a Batch Normalization (left box). Alternatively, a Residual ThriftyNet perform the same operation, as well as a weighted sum of this result with previous iterations before the normalization step (right box). In both cases, the final tensor $x_T$ is fed into a global max pooling, extracting one feature per filter, and into a fully connected layer, connecting it to the output classes. The resulting architecture contains very few parameters mostly determined by the number of feature maps in the convolutional layer.}
    \label{fig:flow_diagram}
\end{figure*}

Many different approaches were explored in the field of neural network compression, always with the goal of finding the best trade-off between resource efficiency and model accuracy. Most of those methods are orthogonal to one another, meaning that they can be used all together on the same model. For example, pruning, quantization and distillation methods could be applied to ThriftyNets for even smaller model size. In the next paragraphs, we introduce the main contributions to the field, grouped by the type of compression they perform.

\paragraph{Pruning}
Pruning, first introduced by~\cite{lecun1990optimal}, aims at deleting parameters, channels or parts of a network while preserving the global performance~\cite{blalock2020state}. While setting single parameters to zero induces sparsity in both the intermediate representations and parameter tensors, deleting complete channels is harder to achieve with good performance, yet allows faster inference time and less resource consumption. Pruning can be performed once and for all after training, in order to reduce the model's size~\cite{lecun1990optimal, li2016pruning, han2015learning, luo2017thinet}, or during training 
\cite{han2015deep, hacene2019attention} to impact training time as well. In the first case, the stress is put on defining a metric for deleting the least useful neurons or channels. In the latter, mechanisms are designed in order to force the network to abandon the use of some of its parameters during training. Pruning can be applied to the proposed ThriftyNets to reduce their number of parameters.

\paragraph{Quantization}

While standard floating point values have 32 bits of precision, many works have experimentally demonstrated that neural networks do not lose a lot of performance when their parameters are restricted to a small set of possible values~\cite{gupta2015deep}, up to binary neural networks with only two possible values and one bit storage for each parameter~\cite{hubara2016binarized}. Reducing precision allows models to be more compact by a great factor, and allows implementation on dedicated low precision hardware~\cite{merolla2014million,farabet2011neuflow}. Like pruning, quantization can be performed after a training through a transformation of the parameters~\cite{gong2014compressing, denton2014exploiting, choi2016towards}, or during training~\cite{courbariaux2015binaryconnect}. Despite the fact quantization can greatly benefit the memory usage of CNNs, it usually does not reduce the number of parameters and is therefore quite different from the aim of this paper.

\paragraph{Distillation}
Distillation techniques consist in training a deep neural network, termed 'student', to reproduce the outputs of another model, termed 'teacher', with the student being typically smaller than the teacher. While initially only considering the final output of the teacher model~\cite{hinton2015distilling}, methods evolved to take into account intermediate representations~\cite{romero2014fitnets, koratana2018lit}. Distilling a model into itself, or self-distillation, has also proven to be effective when iterated \cite{furlanello2018born}.
While individual knowledge distillation focused on the student mimicking the outputs of the teacher, relational knowledge distillation~\cite{park2019relational, lassance2019deep} made it reproduce the same relations and distances between training examples, yielding a better representation of the latent space for the student, and better generalization capabilities.

\paragraph{Efficient scaling}

While the diversity of datasets in computer vision does not cease to increase, some works are focused on the re-usability of architectures that perform well on specific tasks. Mainly, the resolution and complexity of the input image play a great role on the minimal number of parameters required to achieve good performance. While it is possible to scale architectures by adding layers (depth increase)~\cite{he2016deep} or by adding filters (width increase)~\cite{howard2017mobilenets, zagoruyko2016wide}, a correct balance between those hyperparameters has been shown to lead to better results~\cite{tan2019efficientnet}.

\paragraph{Factorization}
Factorization consists in reusing parameters, channels or whole parts of the network several times, thus effectively reducing its size compared to a counterpart where the repeated parts would be made of distinct elements. Parameters can be grouped by values, and accessed through an indirection table~\cite{wu2018deep}, in approaches that are often coupled with quantization~\cite{chen2015compressing}. Convolution kernels of great size can be factorized into smaller kernels before training~\cite{chollet2017xception, iandola2016squeezenet} or using matrix factorization after training~\cite{denton2014exploiting}. The proposed architectures can be though of as heavily factorized CNNs.

\paragraph{Recurrent residual networks as ODE}

Since the initial proposal of residual networks~\cite{he2016deep}, many works have studied them theoretically, observing that the forward pass of a residual network resembles the explicit Euler scheme of an ordinary differential equation~\cite{zhang2019forward, avelin2019neural, chen2018neural}. The question of stability, inversibility and reusability of the convolutional filters became central~\cite{jacobsen2018revnet, behrmann2018invertible}. Experiments were conducted on recurrent residual networks with a single filter iterated over time~\cite{liao2016bridging}. Those studies provide theoretical insight on why reusing filters at different depths can be effective. The proposed method is highly inspired by those works.

\section{Methodology}
\label{sec:methodo}

\subsection{Context}

CNNs form a family of Deep Neural Networks which parameters are mainly arranged in convolutional layers. Such layers are determined by two tensors $\mathbf{W}$ and $\mathbf{B}$, corresponding to the kernels and biases parameters respectively. Denoting by $f_{in}$ the number of input channels, $f_{out}$ the number of output channels, $a$ the kernel width and $b$ the kernel height, the cardinality of $\mathbf{W}$ can be written as $f_{in} f_{out} a b$, while $\mathbf{B}$ has cardinality $f_{out}$.

In most cases, architectures become wider in their deep layers, where the spatial dimension of processed signals is reduced. As a consequence, most of the parameters are contained in the deep layers, while the number of operations is almost evenly distributed along the architecture~\cite{he2016deep}. In an effort to reduce the number of parameters in deep convolutional neural networks, it is therefore usual to target the deep layers in priority. This is in contradiction with the fact that the number of parameters scales quadratically with the depth of the architecture in many models. This is even more problematic since state-of-the-art results often rely on the use of (very) deep neural networks.

In order to remove this dependency, we propose to share kernels between layers, from the input to the output of the considered architecture. Similar ideas were proposed in previous works~\cite{liao2016bridging}. As a result, the proposed architectures can be scaled to any depth with little impact on the total number of trainable parameters.

\subsection{Thrifty Networks}

Consider a problem in which the aim is to associate an input tensor $\mathbf{x}$ with an output $\mathbf{y}$ through the network function $f$. This network function is trained using a variant of the stochastic gradient descent algorithm to minimize a loss function $\mathcal{L}$ over a dataset $\mathcal{D}$. 

We propose to define a convolutional layer $\mathcal{C}$, parametrized by weights $\mathbf{W}$, and to build a deep neural network using only this layer iteratively applied $T$ times on successive latent representations of the input. Note that we do not use a convolution with bias in this work.

This architecture, called \textbf{ThriftyNet}, is then defined by the following recursive sequence:
\begin{equation}
\left\{
    \begin{array}{lll}
         \mathbf{x}_0&=& \texttt{PAD}(\mathbf{x})  \\
         
         \mathbf{x}_{t+1}&=& \mathcal{D}_t\left[\texttt{BN}_t\left(\tx + \sigma(\mathbf{W} \star \tx_t)\right)\right]\\
         
        \mathbf{y} &=& \texttt{FC}(\tx_T)
    \end{array}\;,
\right.
\label{definition}
\end{equation}
where \texttt{PAD} creates extra channels filled with constant values to extend the dimension of $\mathbf{x}$, \texttt{BN} is a batchnorm layer, $\mathcal{D}_t$ is a downsampling operation (typically achieved with strides or pooling) or the identity function, and \texttt{FC} is a final fully connected layer.

Several classical activation functions $\sigma$ were considered in this algorithm. In our case, not only do they break the linearity, they also play an important role in regularizing the norm of the activation tensor. We found that the hyperbolic tangent and ReLU yielded the best results in term of accuracy. However, for the purpose of implementing this algorithm onto a specific hardware, we focused on rectified linear units (ReLU).

Note that convolutions can be applied to inputs of any spatial dimensions, which is why we can reduce the spatial dimension of the inputs throughout the process.

\subsection{Residual Thrifty Networks}

In order to boost the performance of thrifty networks, we add a shortcut mecanism, where activations from previous iterations can be added from the past. This \textbf{residual thrifty network} adds $T (h+1)$ parameters on top of a regular thrifty network, with $h$ being a hyperparameter representing how many steps in history are kept in memory when processing a new iteration. They are grouped in a matrix $\mathbf{\alpha}$. Those parameters are the coefficients weighting the contribution from past activations at each step. In residual thrifty nets, Equation~(\ref{definition}) is replaced as follows:

\begin{equation}
\left\{
    \begin{array}{lll}
    \mathbf{x}_0&=& \texttt{PAD}(\mathbf{x})  \\
         
    \mathbf{a}_{t+1} &=& \sigma(\mathbf{W} \star \tx_t) \\
    \mathbf{b}_{t+1} &=& \displaystyle{\mathcal{D}_t\left[\mathbf{a}_{t+1}\right] + \sum_{i=0}^h{ \alpha[t,i] \displaystyle{\mathop{\bigcirc}_{j=0}^{t}{\mathcal{D}_j\left[\tx_{t-i}\right]}}}}\\
    \tx_{t+1} &=& \texttt{BN}_t\left(\mathbf{b}_{t+1}\right)\\
         
    \mathbf{y} &=& \texttt{FC}(\tx_T)
    \end{array}\;,
    \right.
\end{equation}
where $\bigcirc$ denotes the composition operator. Previous activations $\mathbf{x}_{t-i}$ are added only if $t-i \geqslant 0$.

Adding contributions from the past lead to better performance of the architecture on every tasks at the expanse of only a handful of additional parameters. However, the cost of computation is slightly increased, as well as the memory requirements to store previous activations. This trade-off will be discussed in the experiments section.

\subsection{Pooling strategy}

Pooling has two notables effects: firstly, it diminishes the number of computations made by one iteration which significantly increases the speed of a forward pass in the model. Secondly, it allows the convolutional filter to take effect into larger regions of the initial images.
In the residual thrifty network, pooling is applied to every elements in the history, in order to guarantee size compatibility. The pooling positions are set as hyperparameters, and various strategies are possible and discussed in the experiments.

As a consequence, once the hyperparameters of the convolution are fixed, a ThriftyNet is characterized by an integer sequence $\left(\mathcal{D}_t\right)_{0\leq t < T}$, where $\mathcal{D}_t$ denotes the downsampling that occurs at step $t$ (1 means by convention that no downsampling is performed at this step).

\subsection{Grouped convolutions}

In our models, we found that using grouped convolutions can lead to better performance for some tasks. Recall that a group convolution is obtained by splitting the input tensor along the feature maps axis, and performing computations on each resulting slice concurrently with independent kernels. In more details, in some experiments we design convolutions that are obtained by composing two elementary convolutions: the first one uses a kernel-size $a b$ and as many groups as feature maps in the input. The second one uses kernel-size of 1 and only one group. As a result, the first convolution exploits the spatial structure of the input, but treats each feature maps independently, whereas the second convolution disregards the spatial structure but mixes feature maps. The total number of parameters in the weights of such a convolution is therefore $f_{out} a b + f_{in} f_{out}$.

\subsection{Hyperparameters and size of the model}

In Table~\ref{tab:params_recap} we recap the hyperparameters to define ThriftyNet and their residual version. These hyperparameters are recalled or defined in the following list:
\begin{itemize}
    \item The number of filters $f$ in the convolutional layer
    \item The dimension $(a,b)$ of the convolution kernel
    \item The number of iterations $T$
    \item The number of steps $h$ to consider when performing shortcuts from previous activations
    \item The downsampling sequence $(\mathcal{D}_t)_t$
\end{itemize}

\begin{table}[h]
    \centering
    \begin{tabular}{r c l}
        \toprule
         Model & Convolutions & Size \\
         \midrule
         ThriftyNet & Classical & $f^2ab + 2fT$ \\
         ThriftyNet & Grouped & $f(ab + f) + 2fT$ \\
         Residual ThriftyNet & Classical & $f^2ab + 2fT + hT$ \\
         Residual ThriftyNet & Grouped & $f(ab+f) + 2fT + hT$ \\
         \bottomrule
    \end{tabular}
    \vspace{\baselineskip}
    \caption{Summary of the number of parameters of the proposed models, as functions of the hyperparameters}
    \label{tab:params_recap}
\end{table}

\section{Experiments}
\label{sec:expe}

 In this section, we explore the performances obtained with various values of hyperparameters, namely the total number of iterations $T$, the number of downsamplings performed, the history, and the number of filters $f$, which is directly linked to the number of parameters. 

Experiments are performed on CIFAR-10, CIFAR-100 and SVHN. CIFAR-10 and CIFAR-100~\cite{krizhevsky2009learning} are datasets of tiny colored images of 32x32 pixels. They contain both 50,000 samples for training and 10,000 samples for test. CIFAR-10 is made of 10 classes, whereas CIFAR-100 contains 100 classes. Note that state-of-the-art performance is 98.9\% on CIFAR-10 and 91.70\% on CIFAR-100 using EfficientNet-B7~\cite{tan2019efficientnet} (64M parameters). SVHN is a dataset of digit classification from pictures of house numbers. Images are also of size 32x32 and present up to three digits in them. The classification task consists in identifying the central digit. The state-of-the-art performance without data augmentation was achieved by a Wide-ResNet-16-8~\cite{zagoruyko2016wide} with 98.46\% accuracy.

We use Stochastic Gradient Descent as our optimizer, starting with a learning rate of $10^{-1}$ and dividing it by 10, usually at epochs 50, 100 and 150, for a total of 200 epochs.

\subsection{Impact of data augmentation}

As we work with tiny networks, regularization techniques like heavy data augmentation, mixup or cutmix, designed to help very expressive networks to achieve better generalization performance, have little to no effect, as observed in our experiments. In Table~\ref{tab:data_augment}, we report the performance on both the train and test sets we obtained when using a residual ThriftyNet with 40k parameters and 20 iterations. We observe very little impact in using more advanced techniques of data augmentation. We hypothesize that thrifty architectures are less likely to cause overfit because of the very constrained number of parameters. On SVHN, a 20K parameter ThriftyNet, trained with Auto Augment~\cite{cubuk2018autoaugment} and Cutout of size 8~\cite{devries2017improved} achieves 97,25\% accuracy, compared to the 96,59\% of the same architecture trained on the raw dataset.

\begin{table}[]
    \centering
    \begin{tabular}{rcc}
        \toprule
        Data augment & Test accuracy \\
        \midrule
        Standard (crops and horizontal flips) & 90,64\% \\
        Standard + auto augment~\cite{cubuk2018autoaugment} & 91.00\% \\
        Standard + cutout size 8~\cite{devries2017improved} & 90.40\% \\
        Standard + mixup~\cite{zhang2017mixup} & 88.09\% \\
        Standard + cutmix~\cite{yun2019cutmix} & 88.47\% \\
        \bottomrule\\
    \end{tabular}
    \caption{Impact of the chosen data augmentation technique on the test sets accuracy when training residual ThriftyNets with 40k parameters and 20 iterations.}
    \label{tab:data_augment}
\end{table}

As a result, in the following experiments, we only use standard augmentation consisting of horizontal flips (for CIFAR only) and random crops.

\subsection{Comparison with standard architectures}

We then compare in Table~\ref{tab:results_cifar10} and in Table~\ref{tab:results_cifar100} the performance of the proposed ThriftyNet architectures to standard ones found in the literature, resp. on CIFAR-10 and CIFAR-100. When available, the reported scores are those obtained with standard data augmentation. Otherwise, they are obtained using only the raw training set.

The first observation that we can draw from Table~\ref{tab:results_cifar10} and Table~\ref{tab:results_cifar100} is that ThriftyNets present competitive results with the literature for a tiny parameter budget. The Residual ThriftyNet we use on CIFAR-10 achieves up to 91\% accuracy while using less than 40K parameters in total. The only size-comparable architecture we found is the one introduced in~\cite{liao2016bridging}, but its accuracy is significantly lower than that of the ThriftyNets. Since it is not completely fair to compare architectures that target different accuracies, we run an additional experiment in which we scale down DenseNet-BC and Resnet so that they contain a comparable number of parameters. More precisely, we proportionally reduce the number of feature maps of every single convolutional layer in the considered architectures until reaching the targeted number of parameters. We obtain that DenseNet-BC achieves 87.91\% accuracy, and Resnet 86.72\% accuracy. Again, we observe a significant drop in accuracy compared to ThriftyNets. 

What we draw from these experiments is that when the parameter budget is very constrained, ThriftyNets appear as an interesting choice.
For CIFAR-100, the proposed Residual ThriftyNet architecture achieves 74.37\% accuracy, which is competitive regarding its 600K parameters in total. 

However, let us point out that this performance comes at the expense of the number of operations performed by the network. Since ThriftyNets apply the same convolutional filter at each iteration, first iterations using both the full depth of the filter and the full resolution of images are very costly. We investigate this limitation later in this section.

\begin{table}[h]
  \centering
  \begin{tabular}{rccc}
  \toprule
    Model & Parameters & Macs & Test accuracy\\
    \midrule
    Resnet-110~\cite{he2016deep} & 1.7M & 250M & 93,57\% \\
    FitNet\*~\cite{romero2014fitnets} & 1.6M & - & 91.10\% \\
    SANet~\cite{hacene2019attention} & 980K & <42M & 95,52\% \\
    Pruned ShuffleNets~\cite{paupamah2020quantisation} & 879K & < 4M & 93,05\% \\
    DenseNet-BC~\cite{huang2017densely} & 800K & 129M & 95,49\% \\
    Pruned MobileNet~\cite{paupamah2020quantisation} & 671K & < 7.8M & 91.53\% \\
    Wide-ResNet~\cite{zagoruyko2016wide} & 564K & 84.3M & 93.15\% \\
    IRKNets\*~\cite{zhu2018convolutional} & 320K & - & 92,82\% \\
    Resnet-20~\cite{he2016deep} & 270K & 40M & 91,25\% \\
    3-state Recurrent Resnet~\cite{liao2016bridging} & 121K & - & 92,53\% \\
    Fully Recurrent Resnet~\cite{liao2016bridging} & 39,7K & - & 87\%\\
    \midrule
    Tiny Resnet (ours) & 43.6K & 6.8M & 86.72\%\\
    Tiny DensetNet-BC (ours) & 39,6K & 6M & 87,81\%\\
    \midrule
    ThriftyNet h=5, T=15 (ours)& 39,6K & 130M & 90.15 $\pm$ 0.42\% \\
    ThriftyNet h=5, T=45 (ours)& 39,6K & 300M & 90.95 $\pm$ 0.45\% \\
    \bottomrule \\
  \end{tabular}
  \caption{Comparative results on CIFAR-10. Experiments were performed 5 times and the interval is the observed standard deviation. Models followed by a * are trained without the standard data augmentation (horizontal flips and crops).}
  \label{tab:results_cifar10}
\end{table}

\begin{table}[h]
  \centering
  \begin{tabular}{rccc}
  \toprule
    Model & Parameters & Macs & Test accuracy\\
    \midrule
    FitNet\*~\cite{romero2014fitnets} & 2.5M & - & 64.96\% \\
    ResNet-164~\cite{he2016deep} & 1.7M & 257 M & 75.67\% \\
    IRKNets~\cite{zhu2018convolutional} & 1.4M & - & 79.15\% \\
    SANet~\cite{hacene2019attention} & 1.01M & 42M & 77.39\% \\
    DenseNet-BC~\cite{huang2017densely} & 800K & 129M & 77,73\% \\
    Wide-ResNet~\cite{zagoruyko2016wide} & 600K & 84.3M & 69.11\% \\
    \midrule
    ThriftyNet h=5, T=40 (ours) & 600K & 2740M & 74.37\% \\
    \bottomrule \\
  \end{tabular}
  \caption{Comparative results on CIFAR-100. Models followed by a * are trained without the standard data augmentation (horizontal flips and crops).}
  \label{tab:results_cifar100}
\end{table}

\subsection{Effect of the factorization on the filter usage}

\begin{figure}[h]
\centering
    \includegraphics[width=0.45\textwidth]{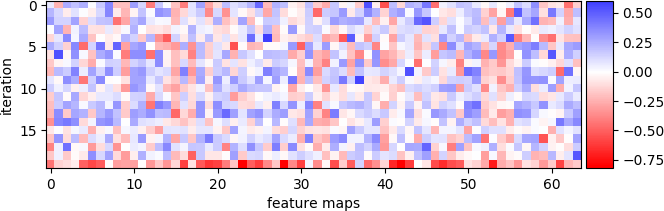}
\caption{Visualization of the mean activation of each convolutional filter at every iterations.}
\label{fig:iter_fm_visu}
\end{figure}

Iterating over the same convolutional filter on a ThriftyNet that presents downsamplings means that the architecture is offered the possibility to reuse filters over different spatial resolution of the inputs. One could then imagine that the optimization scheme would associate some filters with wide spatial representations and some with deeper iterations in the network. In Figure~\ref{fig:iter_fm_visu}, we plot the mean activation of the filters over the training set of CIFAR-10 for each iteration, for a trained Residual ThriftyNet with 64 filters and 20 iterations. We observe that no clear pattern appears in the activation of the filters, from which we deduce that they are being consistently used at every iteration.

\subsection{More computationally efficient ThriftyNets}

In an effort to reduce the impact of the computational cost of the first iterations, it can be beneficial to perform downsamplings sooner than later.
In the last experiments, we used a regular spacing of downsamplings. But here we consider performing them at any iteration. On CIFAR-10, we were able to obtain an accuracy of 85.35\% for a total of 49M Macs (Multiply-accumulate operations), while the regular pooling version achieves a mean of 90.15\% for 130M Macs.

\begin{figure}[h]
    \begin{tikzpicture}
    
    \begin{axis}[
    xlabel={MMacs},
    ylabel={Test accuracy on CIFAR-10 (\%)},
    ylabel near ticks,
    grid=major,
    legend entries={Irregular pooling, Regular pooling},
    legend style={at={(0.99,0.01)}, anchor=south east},
    width=\figurewidth
    ]
    
    \addplot[only marks, color=blue] coordinates {
    (49.37, 85.41)
    (99.97, 90.33)
    (130, 90.05)
    (188.39, 90.11)
    (351.52, 90.05)
    (252.94, 90.05)
    (162.29, 91.08)
    (178.12, 90.83)
    (84.52, 88.0)
    (59.24, 86.93)
    (69.12, 88.26)
    (332.85, 90.87)
    (272.91, 90.77)
    (128.75, 89.92)
    (96.37, 89.18)
    };
    
    \addplot[only marks, color=red] coordinates {
    (94.81, 89.42)
    (94.81, 88.56)
    (94.81, 89.45)
    (94.81, 89.28)
    (94.81, 89.63)
    
    (130, 90.64)
    (130, 89.73)
    (130, 89.76)
    (130, 90.69)
    (130, 89.93)
    
    (171.73, 90.63)
    (171.73, 90.73)
    (171.73, 90.38)
    (171.73, 90.64) 
    (171.73, 91.15)
    
    (232.12, 90.28)
    (232.12, 90.82)
    (232.12, 90.54)
    (232.12, 89.68)
    (232.12, 91.19)
    
    (280.77, 91.59)
    (280.77, 91.34)
    (280.77, 89.91)
    (280.77, 89.89)
    (280.77, 91.44)
    
    (300, 90.76)
    (300, 91.06)
    (300, 90.55)
    (300, 90.58)
    (300, 91.79)
    
    (320, 90.01)
    (320, 90.54)
    (320, 90.83)
    (320, 90.64)
    (320, 91.52)
    };
    
    \end{axis}
    \end{tikzpicture}
    \caption{Accuracy of various ThriftyNets of 40K parameters, in function of the number of Macs they perform. In blue, the points are obtained by considering irregular spacing of downsamplings. In red, the points are obtained by varying the number of iterations, while maintaining a regular spacing between downsamplings.}
    \label{fig:acc_vs_macs}
\end{figure}
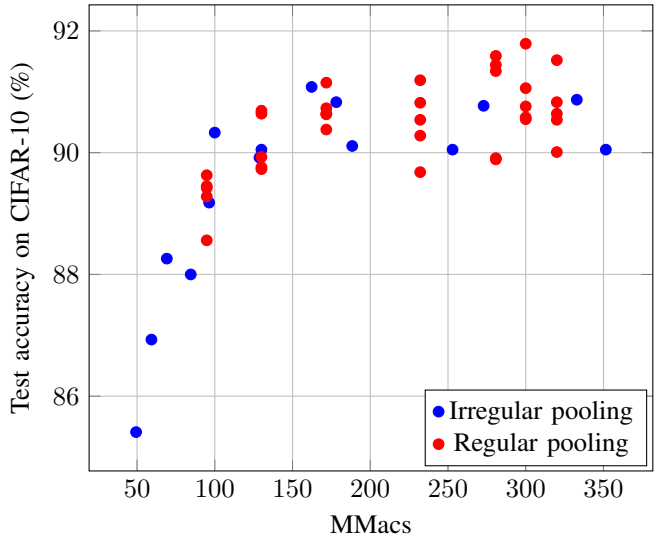

This sheds light on a trade-off in neural network architectures, where a small amount of parameters can be compensated by a large amount of computations. In figure~\ref{fig:acc_vs_macs}, we illustrate this phenomenon by plotting the final test accuracy on CIFAR-10 for various ThriftyNets in function of their Mac count, for a fixed number of parameters (40K). We distinguish two possibilities: in blue we depict what happens when performing irregularly spaced downsamplings and in orange what happens when the total number of iterations varies but downsamplings are regularly spaced.

Interestingly, the two solutions we compare to reduce the number of operations seem to lead to similar behaviors in terms of the trade-off between Macs and accuracy. While irregular downsamplings may lead to more iterations for the same Macs budget, we hypothesize that reducing the number of iterations on high-resolution intermediate representations can lead to significant drops in accuracy.

\subsection{Effect of the number of iterations}

\begin{figure}[h]
    \centering
    \begin{tikzpicture}
    \begin{axis}[
        xlabel={Number of iterations},
        ylabel={Test accuracy on CIFAR-10 (\%)},
        ylabel near ticks,
        ymin=89, ymax=92,
        grid=major,
        legend entries={History 1, History 5, History 10},
        legend style={at={(0.99,0.01)}, anchor=south east},
        width=\figurewidth
    ]
    
    \addplot+[sharp plot] coordinates {
    (10.0, 89.2)
    (15.0, 90.28)
    (20.0, 90.47)
    (25.0, 90.46)
    (30.0, 90.32)
    (35.0, 90.43)
    (40.0, 90.44)
    (45.0, 90.18)
    (50.0, 89.99)};
    
    \addplot+[sharp plot] coordinates {
    (10.0, 89.27)
    (15.0, 90.15)
    (20.0, 90.71)
    (25.0, 90.55)
    (30.0, 90.5)
    (35.0, 90.43)
    (40.0, 90.83)
    (45.0, 90.95)
    (50.0, 90.71)
    };
    
    \end{axis}
    \end{tikzpicture}
    \caption{Accuracy on CIFAR-10 in function of the number of iterations. The number of parameters is fixed at 40K, and pooling is evenly spaced in order to be performed 4 times, plus one last time at the end. Experiment were performed 5 times and the mean accuracy is plotted.}
    \label{fig:iteration}
\end{figure}
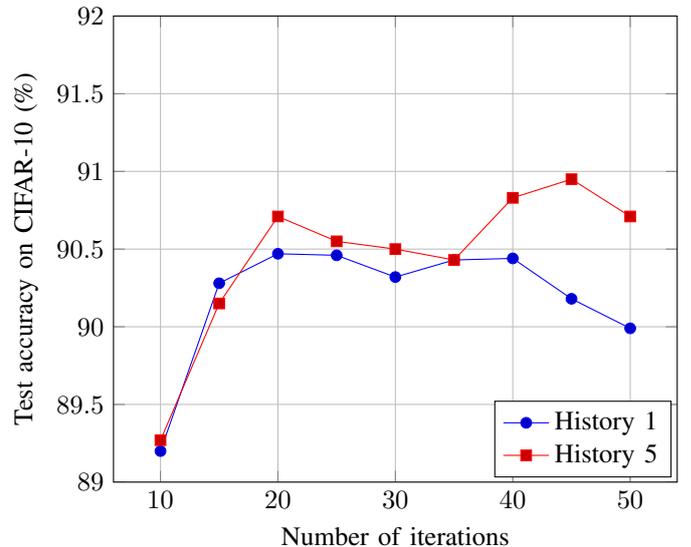

Another important hyperparameter to design ThriftyNets is the number of iterations we use. Normally, this hyperparameter directly influences the number of parameters in the architecture (c.f. Table~\ref{tab:params_recap}). As a consequence, it is difficult to analyze the influence of $T$. In Figure~\ref{fig:iteration}, we report the test-set accuracy of ThriftyNets and residual ThriftyNets when varying the number of iterations $T$, but for a fixed parameters budget of 40k that we reach by tuning the number of feature maps $f$ accordingly. 

First, let us point out that extreme values of $T$ would necessarily lead to poor performance. Indeed, choosing $T = 1$ would lead to a shallow network with poor generalization abilities. Choosing a too large $T$ would cause the number of filters to become too small to expect good performance. What we observe in the experiments is that the choice of $T$ could be more tightly restricted since in the case of ThriftyNets, $T=10$ and $T=50$ lead to poorer performance than values in between. Yet, in the range $[20..40]$, we observe that tuning the number of iterations for a given parameters budget has little influence on overall performance.

\subsection{Effect of the number of filters}

Next, we investigate the evolution in accuracy when the number of filters $f$ increases. Of course, increasing the number of filters leads to more parameters, as shown in Table~\ref{tab:params_recap}. Figure~\ref{fig:n_params} shows the evolution of accuracy for CIFAR-10 and SVHN in function of the number of parameters, obtained when varying $f$. In both cases, we observe that the trade-off between number of parameters and accuracy is not linear: reaching 1\% extra accuracy can be very costly in terms of the required number of parameters, if the accuracy is already high. This trade-off is even sharper in the case of SVHN where we observe a two-step phenomenon, where accuracy is first increased quickly but then saturates. We think that this is a consequence of the difficulty of considered datasets: achieving 93\% accuracy on CIFAR-10 is significantly harder than in the case of SVHN.

\begin{figure*}[h]
  \centering
  \begin{tabular}{cc}
  
  \begin{tikzpicture}
  \begin{axis}[
    xlabel={Number of parameters},
    ylabel={Test accuracy on CIFAR-10 (\%)},
    ylabel near ticks,
    grid=major,
    legend entries={20 Iterations, 30 Iterations, 40 Iterations, resnet},
    legend style={at={(0.99,0.01)}, anchor=south east},
    width=\figurewidth
  ]
    \addplot+[sharp plot] coordinates {
    (3042, 75.05)
    (8002, 84.45)
    (15010, 86.95)
    (24066, 89.27)
    (35170, 90.07)
    (48322, 90.57)
    (80770, 91.58)};
     
    \addplot+[sharp plot] coordinates {
    (3742, 78.64)
    (9342, 85.54)
    (16990, 88.42)
    (26686, 89.68)
    (38430, 90.72)
    (52222, 91.03)
    (85950, 92.19)};
    
    \addplot+[sharp plot, color=green] coordinates {
    (4442, 75.79)
    (10682, 85.39)
    (18970, 88.90)
    (29306, 89.82)
    (41690, 90.04)
    (56122, 91.38)
    (91130, 91.62)};
    \end{axis}
    \end{tikzpicture}

&
 
  \begin{tikzpicture}
  \begin{axis}[
    xlabel={Number of parameters},
    ylabel={Test accuracy on SVHN (\%)},
    ylabel near ticks,
    grid=major,
    legend entries={20 Iterations, 30 Iterations, 40 Iterations},
    legend style={at={(0.99,0.01)}, anchor=south east},
    width=\figurewidth
  ]
    \addplot+[sharp plot] coordinates {
    (1912.0, 87.57)
    (4960.0, 95.17)
    (7816.0, 96.05)
    (9972.0, 96.19)
    (14760.0, 96.27)
    (19852.0, 96.63)
    (39862.0, 96.69)
    };
     
    \addplot+[sharp plot] coordinates {
    (1936.0, 87.15)
    (4950.0, 93.3)
    (7942.0, 95.92)
    (9972.0, 96.13)
    (14886.0, 96.36)
    (19800.0, 96.77)
    (39636.0, 96.81)
    };
    
    \addplot+[sharp plot, color=green] coordinates {
    (1960.0, 85.93)
    (4940.0, 92.2)
    (7900.0, 95.93)
    (9790.0, 95.98)
    (14830.0, 96.33)
    (19852.0, 96.59)
    (39620.0, 96.98)
    };
    
    \end{axis}
    \end{tikzpicture}
\end{tabular}
    \caption{Accuracy on CIFAR-10 (left) and SVHN (right) in function of the number of parameters. Experiments were performed for different numbers of iterations (T) of a Residual ThriftyNet. $h$=5. Pooling was performed every T/5 iterations. The number of filters varied from 32 to 256 by increments of 32.}
    \label{fig:n_params}
\end{figure*}

\subsection{Effect of the number of downsamplings}

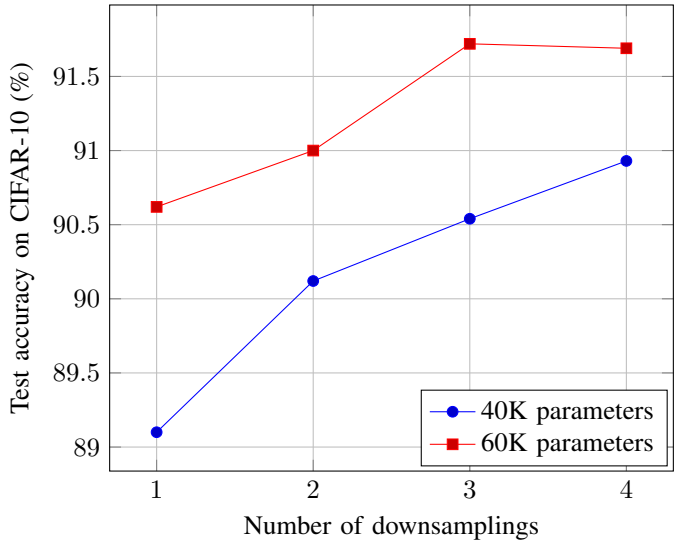
\begin{figure}[h]
    \centering
    \begin{tikzpicture}
    \begin{axis}[
        xlabel={Number of downsamplings},
        ylabel={Test accuracy on CIFAR-10 (\%)},
        ylabel near ticks,
        grid=major,
        xtick={1,2,3,4},
        legend entries={40K parameters, 60K parameters},
        legend style={at={(0.99,0.01)}, anchor=south east},
        width=\figurewidth
    ]
    
    \addplot+[sharp plot] coordinates {
    (1.0, 89.1)
    (2.0, 90.12)
    (3.0, 90.54)
    (4.0, 90.93)
    };
    
        \addplot+[sharp plot] coordinates {
    (1.0, 90.62)
    (2.0, 91.0)
    (3.0, 91.72)
    (4.0, 91.69)
    };
    
    \end{axis}
    \end{tikzpicture}
    \caption{Accuracy on CIFAR-10 in function of the number of downsamplings. 30 iterations are performed, and downsamplings are evenly spaced, with a global max pooling at the very end.}
    \label{fig:ndownsampling}
\end{figure}

Then, we perform an ablation experiment for investigating the importance of pooling in our architecture. We fix the number of parameters and number of iterations, and we report in Figure~\ref{fig:ndownsampling} the evolution of the accuracy as a function of the number of downsamplings. Without surprise, we observe that the more downsamplings, the better the accuracy.

Interestingly, increasing the number of downsamplings has a decreasing consequence on the number of operations that are performed. So contrary to what we reported in Figure~\ref{fig:acc_vs_macs}, here increasing the number of downsamplings is beneficial to both computational complexity and accuracy.

\subsection{Fixing the shortcut parameters in a residual ThriftyNet}

In most modern architectures, shortcut mechanisms consist in adding previous activations to the current one, thus bypassing some of the layers. While they are often fixed and involve only one past activation, we designed residual ThriftyNets to take into account the $h$ last activations, weighted by parameters $\alpha$. This is designed with the hope that the optimization of $\alpha$ leads ThriftyNet into finding the most efficient architectures, and avoid the introduction of additionnal hyperparameters and user priors.

To demonstrate this phenomenon, we perform an ablation experiment. We train a Residual ThriftyNet with an additional loss $\mathcal{L}_\alpha$, designed to make the shortcut parameters converge to 0 or 1. More precisely: $$\mathcal{L}_\alpha = \lambda \sum_{x \in \alpha}{x^2(1-x)^2}.$$ 

We perform 150 epochs of training, with $\lambda$ being multiplied by $1+\epsilon$ after every forward and backward pass of a batch (500 batches per epoch, $\lambda = 3.10^{-4}$, $\epsilon = 1.5.10^{-4}$). Parameters $\alpha$ are then binarized using a threshold at 0.5. From this, we train for 150 additional epochs:

\begin{enumerate}[label=(\alph*)]
    \item The same model without resetting the other parameters
    \item The same model starting from the same initialization
    \item The same model starting from another (random) initialization
\end{enumerate}

Table~\ref{tab:fixed_alpha} sums up the results obtained on CIFAR-10 for this experiment. We observe that the baseline gives the best results. This was expected since shortcuts remain free parameters that can take values others than 0 or 1. Once shortcuts have been fixed and binarized, training from the same initialization is what ranks next. In our experiments, it evens outperforms fine tuning, as the binarization step has a dramatic effect on the accuracy. Training from scratch with a random initialization and the same shortcuts leads to a drop of about 1\% accuracy.

\begin{table}[h]
    \centering
    \begin{tabular}{ r c}
    \toprule
        Model & Test Accuracy  \\
        \midrule
        Baseline accuracy & 91.08\% \\
        After binarization and fine tuning (a) & 88.50\% \\
        After training from the same initialization (b) & 90,47\% \\
        After training from another initialization (c) & 89.98\% \\ 
    \bottomrule \\
    \end{tabular}
    \caption{Shortcut freezing experiment on CIFAR-10.}
    \label{tab:fixed_alpha}
\end{table}

\section{Conclusion}
\label{sec:conclusion}

We introduced ThriftyNet, a convolutional neural network architecture that explores the limits of layer factorization and the efficacy of architectures with tiny parameter count. Based around a single convolutional layer, ThriftyNets iterate over this layer, alternating convolution operations with non-linear activation, batch normalization, downsampling through pooling operations and weighted sums with results from previous iterations. This leads to a very compact architecture that achieves competitive results regarding the trade-off between total number of parameters and accuracy. Such a solution would be beneficial to memory-constrained systems. 
In future work, we consider investigating other strategies to mitigate the large computational cost of ThriftyNets. 

\bibliographystyle{unsrt}  
\bibliography{references}

\end{document}